# High Throughput Multi-Channel Parallelized Diffraction Convolutional Neural Network Accelerator

*Zibo Hu[1], Shurui Li[2], Russell L.T. Schwartz[1], Maria Solyanik-Gorgone[1], Mario Miscuglio[1], Puneet Gupta[2], Volker J. Sorger[1,]\**

[1]**Department of Electrical and Computer Engineering, George Washington University, Washington DC, 20052, USA.**
[2]**Department of Electrical and Computer Engineering, University of California, Los Angeles, California 90095, USA.**
*Corresponding Author Email: sorger@gwu.edu

**Abstract:** Convolutional neural networks are paramount in image and signal processing including the relevant classification and training tasks alike and constitute for the majority of machine learning compute demand today. With convolution operations being computationally intensive, next generation hardware accelerators need to offer parallelization and algorithmic-hardware homomorphism. Fortunately, diffractive display optics is capable of million-channel parallel data processing at low latency, however, thus far only showed tens of Hertz slow single image and kernel capability, thereby significantly underdelivering from its performance potential. Here, we demonstrate an operation-parallelized high-throughput Fourier optic convolutional neural network accelerator. For the first time simultaneously processing of multiple kernels in Fourier domain enabled by optical diffraction has been achieved alongside with already conventional in the field input parallelism. Additionally, we show an about one hundred times system speed up over existing optical diffraction-based processors and this demonstration rivals' performance of modern electronic solutions. Therefore, this system is capable of processing large-scale matrices about ten times faster than state of art electronic systems.

## 1. Introduction:

One of the breakthroughs in artificial intelligence (AI) is associated with convolutional neural networks (CNN) capable of outperforming the human brain in image classification accuracy [1]. Since this discovery, the field has flourished, enriched by novel virtual data processing tasks such as machine vision and autonomous driving, smart target tracing and face recognition, autopilot and surveillance [2]. However, with feature extraction remaining the fundamental task of CNNs, the computing hardware remains a bottleneck, limited by practical performance metric [3]. This often pushes industry towards suboptimal, yet realistic solutions; for example, already outdated Alexnet consumes about 80% of computational resources to process convolution operations [4]. Despite the complexity of the task, there has been considerable progress in this field, fueled by an availability of massive image datasets and state-of-the-art massively parallel application-specific computer hardware such as Graphics Processing Units (GPU), Tensor Processing Units (TPU) etc. Nonetheless, even with the recent advances in electronic hardware, decision-making system latency stays at the level of tens of milliseconds to implement and process convolutions, while being a relatively power-hungry solution (~400W to drive a GPU in a standard scenario).

These aforementioned developments, paired with the increasingly growing fraction of optical to electronic interconnects in modern telecommunication systems, fully justify apparent interest in integration of optics and electronics on the level of hardware architecture. Photonic NNs have proved themselves as fast and energy efficient due to its integration capability, however, they lack massive parallelism of free space optical accelerator approaches [5-10]; that is, the to-be-processed information (i.e. the data signal) is propagating in the same two-dimensional plane as the chip itself, while in optical free-space systems, that data uses the orthogonal third dimension normal to the programmable circuit plane. This added degree-of-(dimensional)-freedom is used in featured here CNN accelerator offering massive-parallelism, image batch-processing, and incidentally, continuous data streaming capability. Unlike ASIC electric/optical processors that optimize their computing resources by boosting the architecture complexity, optical processors harvest innate properties of coherent laser light which results in focusing and filtering effects that enable *passive* Fourier Transform (FT) and simple tensor algebra calculations [11-17]. However, domain-crossing-based latency overhead, or the unavailability of conventional machine learning tools such as pruning or backpropagation to be seamlessly executed in the optical domain, may generate a fair amount of skepticism in the computer science community. In spite of such practical challenges [18,19], the unrealized potential can be harvested from natural and cheap or free averaging, FT, and stochasticity of laser photon sources. A number of recent material breakthroughs suggest possibilities for domain-crossing speedup and optimization in optical systems , e.g. [20-24].

Indeed, mixed-signal machine learning acceleration, including electronic-optic hybrid systems bear promise due to the low latency and energy efficiency. Though it is compelling to design, build, and test an electronic simulator and hardware of such hybrid accelerators, the idea to combat the inconsistencies with brute force theoretical modelling has produced deeper insights into system performance limits and co-design options [25-28]. However, actual hardware-in-the-loop demonstrations of benchmarking achievable performance of hybrid or mixed-signal AI accelerators is in its infancy, yet progress is being made and that constitutes motivation and aim of this work [29-34].

Here we introduce an electronic-optic Diffraction Convolutional Neural Network (D-CNN) accelerator capable of image batch processing and continuous data-streaming operation. The system harnesses an energy-passive realization of the convolution theorem by deploying a coherent Fourier optical system. This allows reducing the quadratic scaling law of performing convolutions to a non-iterative $\boldsymbol{O}(N^2)$ approach as compared to $\boldsymbol{O}(N\times Log(N))$ scaling of a Fast Fourier Transform (FFT) in electronics and provided the data size is less than two million data-points. Compared to the other Fourier Optical systems[13,14], we firstly show that processing 10's of images in time-parallel is possible, and furthermore demonstrate the capability of multi-kernel parallelism by utilizing the nature diffraction orders, thus leading together to a 100× throughput speed-up over state-of-art systems [14]. While input parallelization has been previously achieved [13], presented here for the first time kernel parallelization is a novel venue for harvesting diffractive nature of light as opposed to treating it as a parasitic source of noise. Such kernel parallelization allowed us to double the efficiency of the setup, which can be further improved up to a factor of eight. Lastly, we achieve about 50×lower latency CNN processing as compared to GPU for an input of a similar matrix dimension and demonstrate the first continuous data streaming by developing an end-to-end AI software-hardware system by combining FPGA technology with the D-CNN accelerator. We test the classifier with three conventionally used for NN systems' benchmarking datasets: MNIST, Google quickdraw and CIFAR-10.

## 2. Results

*2.1 Fourier Convolutional Neural Network*

Fourier optics holds a promise as a computational engine due to its low latency fundamentally limited only by the speed of light. Inherently high data parallelism offers millions of time-parallel channels to be processed and is limited by diffraction effects. Fundamentally [25], we know that the Fresnel diffraction pattern can be approximated as a FT of the input transparency, however, provided a few underlying assumptions including paraxial limitations, monochromaticity of the electromagnetic field, and non-divergence of an optical beam. Even assuming an ideal spatial alignment of the optical part of the D-CNN, there may be a difference between the exact FT, and the result obtained using this hybrid signal accelerator given effects such as pixel blur, lens non-ideality, altered intensity, or phase distributions in the Fourier domain. Novel alignment techniques paired with capabilities of heterogeneous computing systems for compensation for such non-idealities share the merit for successful implementations of optical D-CNNs.

In signal processing, the convolution of two signals is equivalent to the dot-product multiplication in the Fourier domain (convolution theorem). Even aside any optical imperfections, there is a key advantage of exploiting the convolution theorem to reduce the computational complexity in a CNN algorithm. However, aside from these acceleration techniques, the FT remains a major contributor to higher than practical latency of electronic implementations of convolution, from millisecond to nanosecond per single FT operation. This is where optical solutions could offer a passive paraxially limited FT, and simplified dot-product multiplication. Remarkably, supplied with enough optical resolution, the computational complexity of an optic-based convolutional classifier is unity independent of the size of an input and a kernel. In the optical convolution module of such classifier **Figure. 1d,** both the FT and the dot-product multiplication are performed passively by the thin lens and the DMD surface correspondingly (not including training and generating the kernels). To reinforce this point, we consider a two-dimensional (2D) data input and a kernel of the same size in the case of a convolution in the spatial domain (second column, Figure. 1a), an electronically implemented convolution in the Fourier domain (third column, Figure. 1a), and an optical convolution module (fourth column, Figure. 1a).

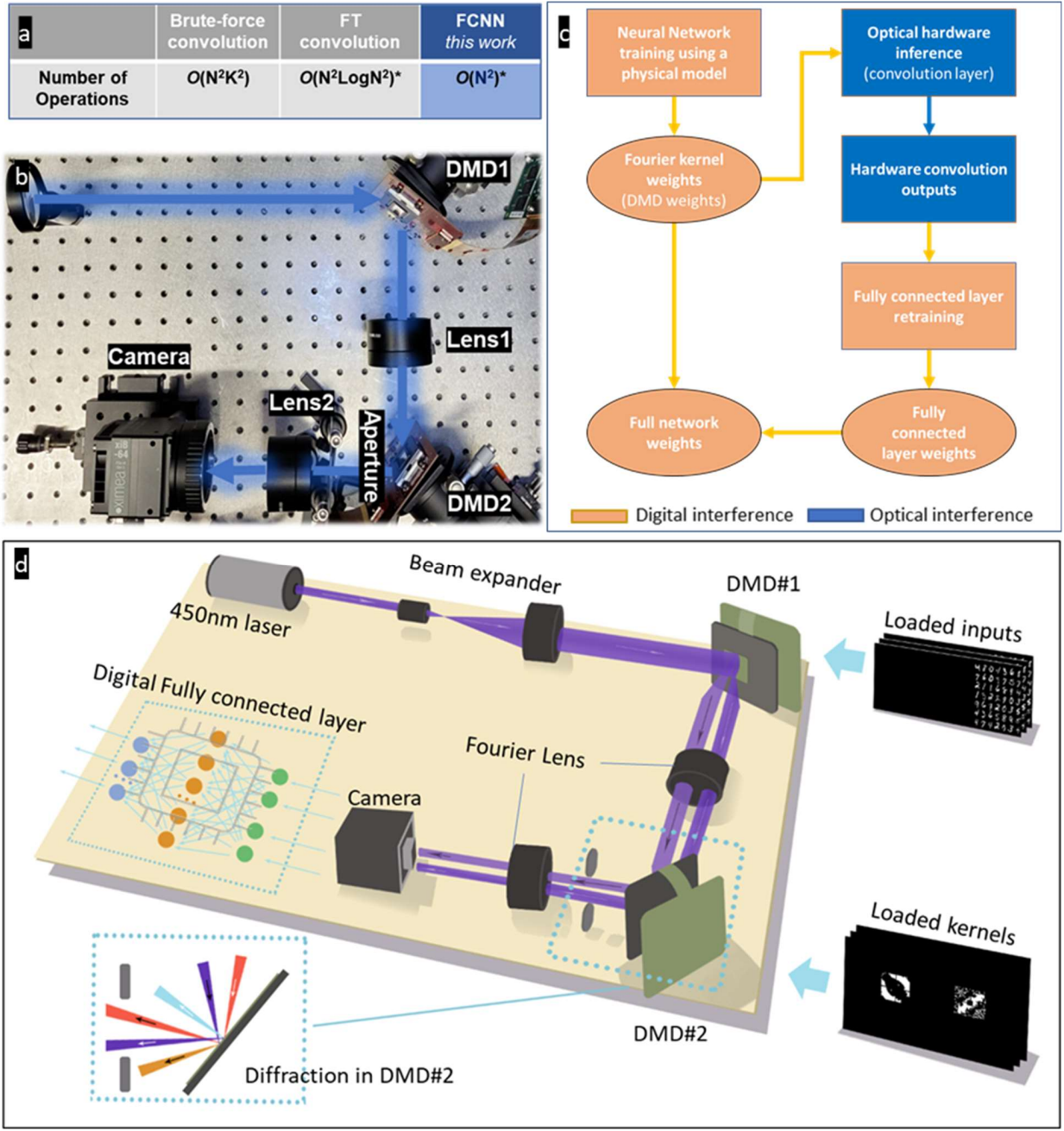

**Figure 1:** Fourier Convolutional Neural Network classifier based on a convolution theorem acceleration. a. The computational operations comparison between our system and an equivalent electric system. The processed image data is N×N size with K×K kernel. *The kernels are trained in the Fourier domain. b. The experimental setup assembled from two 100cm focal length lenses, two DLP LightCrafter 6500 DMDs, and XIMEA-CB019MG-LX-X8G3 camera. c. The system's training flow is a two-stage process; initial stage is training in a GPU-based simulation followed by the second stage of calibration through the D-CNN hardware, to be subsequently used in the classifier. d. Image data is loaded into the optical convolution layer via digital mirror display DMD#1 performing an amplitude signal modulation, Fourier transformed with lens L1, and pixel-by-pixel multiplied in the Fourier domain with the pre-trained set of kernels loaded onto DMD#2. The output is captured by the camera and loaded into the electronic (digital) predictor. The camera performs a non-linear activation alongside with the optical-to-electric domain conversion.

The architecture of the hybrid Fourier classifier is composed of the optical Fourier convolution layer which conducts Fourier domain multiplication by using two lenses in the coherent optical system, followed by an electronic max-pooling layer, and two electronic fully connected (FC) layers including a 256-neurons hidden layer with ReLU activation function and 10-neurons classification layer (Figure.1d). First the electronic NN is pre-trained to generate a set of kernels for a particular dataset. Then, in the optical convolution layer, the system is fed the same dataset of images loaded on DMD#1, and the pre-trained kernels loaded on DMD#2 in the Fourier domain (Figure.1b, d). This converts the costly electronic convolution into a computationally 'free' optical operation. The output is registered by a camera to accomplish the optical-to-electronic domain conversion. Simultaneously, the camera conveniently performs a nonlinear activation due to the intrinsic space-time averaging and squaring of an amplitude of a receiving signal. Upon completion of the (optical) convolution and the domain crossing, the signal is electronically post-processed by two fully connected layers with the ReLU activation function to conclude the classification.

Even though the electronic training model already considers some nonidealities of an experimental setup, the output of the optical layer still differs from that of the hardware due to such factors as alignment noise and electro-optical domain conversion effects, which are challenging to model. The difference between the simulation and the hardware outputs makes the fully connected layer's weights, trained using the simulation outputs, not accurate for hardware outputs. Since the shape of hardware outputs still matches the simulation output, the Fourier kernel weights learned from the simulation model are applicable and the hardware convolution outputs have the same representative power as the simulation model. However, the fully connected layer's weights need to be tuned for the hardware outputs to get maximal classification accuracy. Therefore, we implemented a fine-tuning step, prior to performing the machine learning test step, which uses the hardware convolution outputs to retrain the fully connected layer's weights to the listed in Table 1 experimental accuracy.

*2.2 Harvesting diffraction for multi-kernel parallelism*

Realizing that the DMD generates a diffraction pattern coming from the periodicity ofits 2D-array of micromirrors, we take advantage of multiple copies being generated by the system "for free". This allows us to couple each copy independently to non-identical kernels in the Fourier domain of the first lens. For the first time to our knowledge, we explore paralleling this D-CNN accelerator by performing multiple kernel convolution operations in time-parallel (**Figure. 2b,c**). Such expansion of parallelism allowed us to achieve higher DMD footprint and laser power utilization and resulted in a dramatic expansion of throughput and increase in processing efficiency.

In detail, the diffraction angle is proportional to the wavelength of light and scales with the DMD pitch $\Delta$ such as,

$$\theta = \frac{\lambda}{\Delta} \quad (4)$$

Thus, we place the pre-trained distinctive Fourier kernels at the centers of $0^{th}$ and $1^{st}$ diffraction orders from the input DMD#1 thereby achieving parallelization in Fourier domain (Figure.2a,c). The resulting optical signal is captured by the CCD camera.

While diffraction from DMD#1 is harvested for parallelization, diffraction from DMD#2

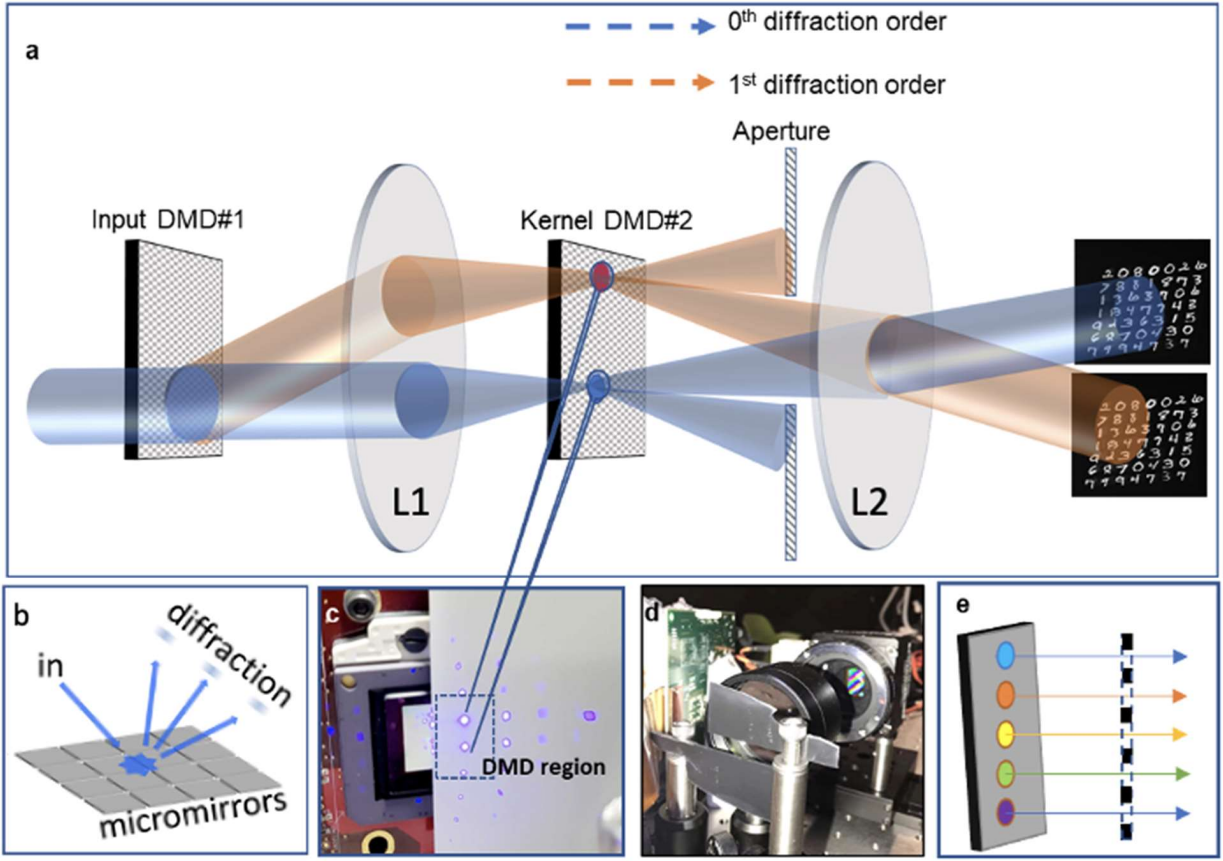

**Figure 2:** Parallel-kernel implementation of the amplitude only 4F system a. The incident laser beam illuminates the DMD#1 which generates different orders in the rear focal plane of L1. We use 0th and 1st orders marked as blue and orange are projected onto DMD#2 in the rear focal plane of L1. Two different kernels are projected onto the DMD#2 in the corresponding regions. The resulting modulated signal leaving DMD#2 also contains multiple diffraction orders which we block with the custom aperture. In the current system, we show 49 inputs x 2 kernels parallelism. b. The schematic representation of the diffraction effects coming from the grid of micromirrors forming a DMD surface. c. A typical diffraction pattern from a DMD. The setup limits the kernel parallelization to two due to the diffraction angle and the focal distance, see Methods. We will further increase the kernel parallelization by changing the wavelength, focal length and/or updating the hardware. D. The aperture blocks parasitic diffraction from DMD#2, which allows the clear output in image plane. E. In a more than 2 kernels system, a multi-kernel custom aperture can be realized by using 3D printing, lithography etc. which filters out the higher diffraction orders from other kernels.

is parasitic and needs to be filtered out. To overcome this effect, we place an adjustable aperture between DMD#2 and the camera to block the diffraction noise, Figure. 2d. In the current setup, we only have 2 channels so we can easily block the unwanted orders. To further increase the parallelism in the Fourier domain, we can design an advanced custom aperture to accomplish filtering for multi-kernel processing with a number of kernels larger than two, Figure. 2e.

## 2.3 Convert 8-bits image into binary DMD inputs

The data input (image) quantization has been applied to pre-process the datasets for the binary modulation (see Methods). To identify the optimal quantization parameters to effectively display an image in the binary DMD we swapped the quantization levels versus the similarity index between the original and recombined images in the simulation. The classification accuracy grows logarithmically as the quantization step increases, which is expected since more information is retrained. However, the tradeoff is the optical classifier performance latency scaling as the number of sub-images grows. We find that a 4-step quantization is the elbow point, which means that when further quantizing the image the increase in structural similarity becomes insignificant. However, latency tradeoff is still expected to grow linearly. Hence, further increase in the number of quantization steps is experimentally unjustified. **(Figure 3)**

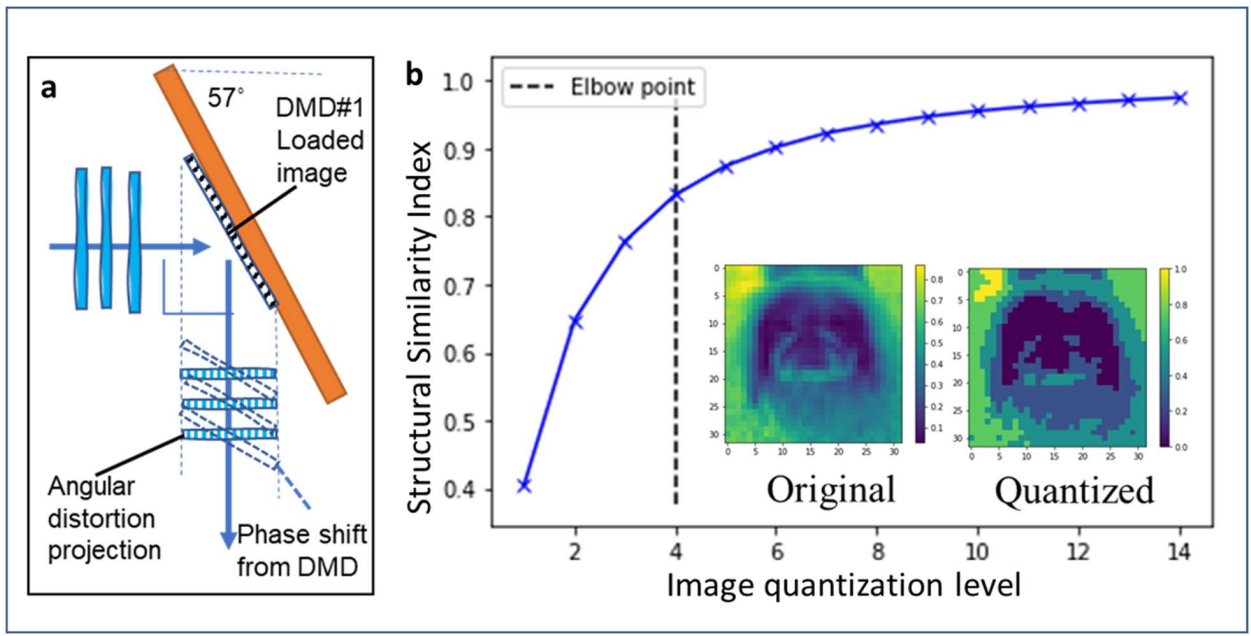

**Figure 3:** Input-data bit resolution impact on D-CNN classifier machine learning accuracy: a. A DMD has two states, the "on" and "off" built-in angles generate a phase shift and angular distortion. We upload pre-processed accordingly distorted images and calibrate the digital weight to compensate for the phase effects; b, The DMD only provides binary amplitude modulation. For the CIFAR-10 8-bits images, we quantize the image with 4 levels effectively expanding the data set by a factor of 12 (4 levels per each of three GRB channels), see Methods for details. We identify an optimal quantization as an elbow point assessing the pre-processed image structural similarity index (SSIM) and looking for an adequate trade-off between system latency and neural network accuracy.

*2.4 Multiple inputs*

We optimized the classifier to enhance the benefits provided by the intrinsic parallelism and fully exploit the nominal bandwidth. The input to the optical convolution layer has been parallelized so that the 5×5 matrix of CIFAR-10 or 7×7 matrix of MNIST/Google quickdraw images have been fed into the classifier. To avoid the crosstalk in the frequency domain, we spatially separate the input images with 30 pixels gap to restrict the crosstalk frequency from overlapping with the significant information carrying frequencies. The high-pass filter is applied to filter out these crosstalk frequencies and increase the misalignment tolerance. In training, this results in applying a mask on the learned Fourier kernel that sets the center 3×3 pixels to zero.

One can see a clear potential to further increase in data throughput by expanding the parallel input beyond 7×7, and the parallel kernel beyond 2. However, limited by the size of the Fourier domain and diffraction limit, one would do so at expense of resolution and classification accuracy. We analyzed the physical limitations on kernel size by sweeping the resolution and estimating the corresponding classification accuracy in our simulation. We found that the smaller the input matrix, the higher is the kernel resolution, and the better is the accuracy, as expected. However, this dependence becomes saturated at 2× kernel resolution in the simulation, and 1× resolution in the experiment with respect to the original size. For example, given the original size of an MNIST image being 28×28 pixels, the accuracy will be saturated at 56×56 kernel size in the simulation, and at 28×28 experimentally. This means that the FT'd image size projected on DMD#2 is not a bottleneck of this D-CNN prototype. A larger input image provides for a more robust output; hence we implement 3× input enlargement to mitigate the DMD#2 alignment difficulty yet giving up throughput.

*2.5 Neural Network accuracy*

The experimental accuracy is consistent with the simulated accuracy in MNIST and Google Quickdraw. The D-CNN system, as expected, shows a solid performance for the datasets which can be converted into binary images with high SSIM **(Table 1)**. However, the experimental accuracy for CIFAR-10 drops by about 10% in the simulation due to the lossy bit quantization procedure, see Methods section. Compared to our previous single input and single kernel Gen 1.0 version [13], the batch processing Gen 1.2 version increased the parallelism from 1×1 to 49×2 with a limited trade-off of 6% accuracy with the CIFAR-10 dataset due to the decreased misalignment tolerance as a tradeoff for the system's form factor and higher parallelism

| Dataset | Simulated accuracy | Experimental accuracy |
|---|---|---|
| MNIST | 99% | 98% |
| Google quickdraw | 93% | 93% |
| CIFAR-10 | 54% | 45% |
| CIFAR-10 without high-pass filter | 65% | 35% |
| CIFAR-10 without parallelism from Gen 1.0 | 65% | 54% |

**Table 1.** Optical neural network performance with MNIST, Google Quickdraw and CIFAR-10. We compare three cases: (1) the optical classifier with high-pass filter; (2) the optical classifier without the high-pass filter and (3) Gen 1.0 with single input single kernel [13]. In the current setup, we trade the accuracy for the batch process ability and parallelization. The GitHub link for our simulation part: https://github.com/nanocad-lab/Optical-NN

## 3. Discussion

The D-CNN has four development stages denoted as Gen1.0 through Gen1.3. Gen 1.0 is the previous non-streaming single-channel generation [14], where we use 300mm focal length lenses. Gen 1.1 and 1.2 are new streaming versions, where 1.1 is multiple inputs with a single kernel and 1.2 is multiple inputs and two kernels, which results in higher data throughput. Considering the system design based on Fresnel propagation, the paraxial approximation, we select 30mm focal length as our projected Gen 1.3 version to further increase the kernel parallelism from 2 to 24 and implement an FPGA to increase the streaming speed from 60Hz HDMI bottlenecked interconnect to 2kHz PCIe bandwidth speed for binarized image processing. (There is a detailed analysis in supplementary S5). In this state-of-the-art system, we assume we had the best COTS devices and solve all connection issues to harvest the 15kHz best DMD performance.

The accuracy of the current multi-channel version dropped from the previous design Gen 1.0. It is caused by two reasons: (1) the denser Fourier domain resulted from a shorter focal length increased the alignment difficulty, and (2) the enlarged numerical aperture departs the system from the paraxial regime resulting into a minimal parasitic phase noise in the Fourier domain. However, we dynamically correct for it by adjusting the physical size of the image pixel as it gets displayed on a DMDs superpixel. This adjusts the maximum spatial frequency of an input image and, hence, relaxes the diffraction limits. Our system's reconfigurability allows such a real-time optimization to selectively optimize for the throughput or for accuracy.

Next, we are interested in determining computational performance metrics of batch-processing and data-streaming with this D-CNN prototype and compare it to the state-of-the-art GPUs. Performance metrics of interest include throughput/efficiency in units of Operations Per Second per Watt (OPS/W) and system latency. In this context the word 'system' refers to the complete stand-alone electronic-optical system and not only to its optical portion. For the cited performance of reference GPUs V100 and M40 on VGG-10, please, see the reference [37]. Theoretical (nominal) Tesla V100 performance is taken from the device datasheet. All the points are based on two equivalent metrics, the brute-force convolution and Fourier convolution (Methods section, **Figure 4**).

Latency is another important parameter from a system performance perspective, and, specifically, for identifying the bottlenecks. Currently, we have found the CCD camera interface to be a major system bottleneck, due to the methods of using the camera and the data processing that image information must undergo to be used by the fully connected layers down the stream. In the present implementation, the camera captures pictures as needed to operate the system, however the speed is limited below the camera's maximum capture speed of 2 kHz. The set up allows for the camera to capture frames at 100 Hz due to the need to interface between code languages and the relatively slow speed of python-based software that the fully connected layer is implemented in. Both issues can be solved by implementing the fully connected layer in the camera's native language of C++, which should allow for optimal software integration and realization of nominal data processing speed.

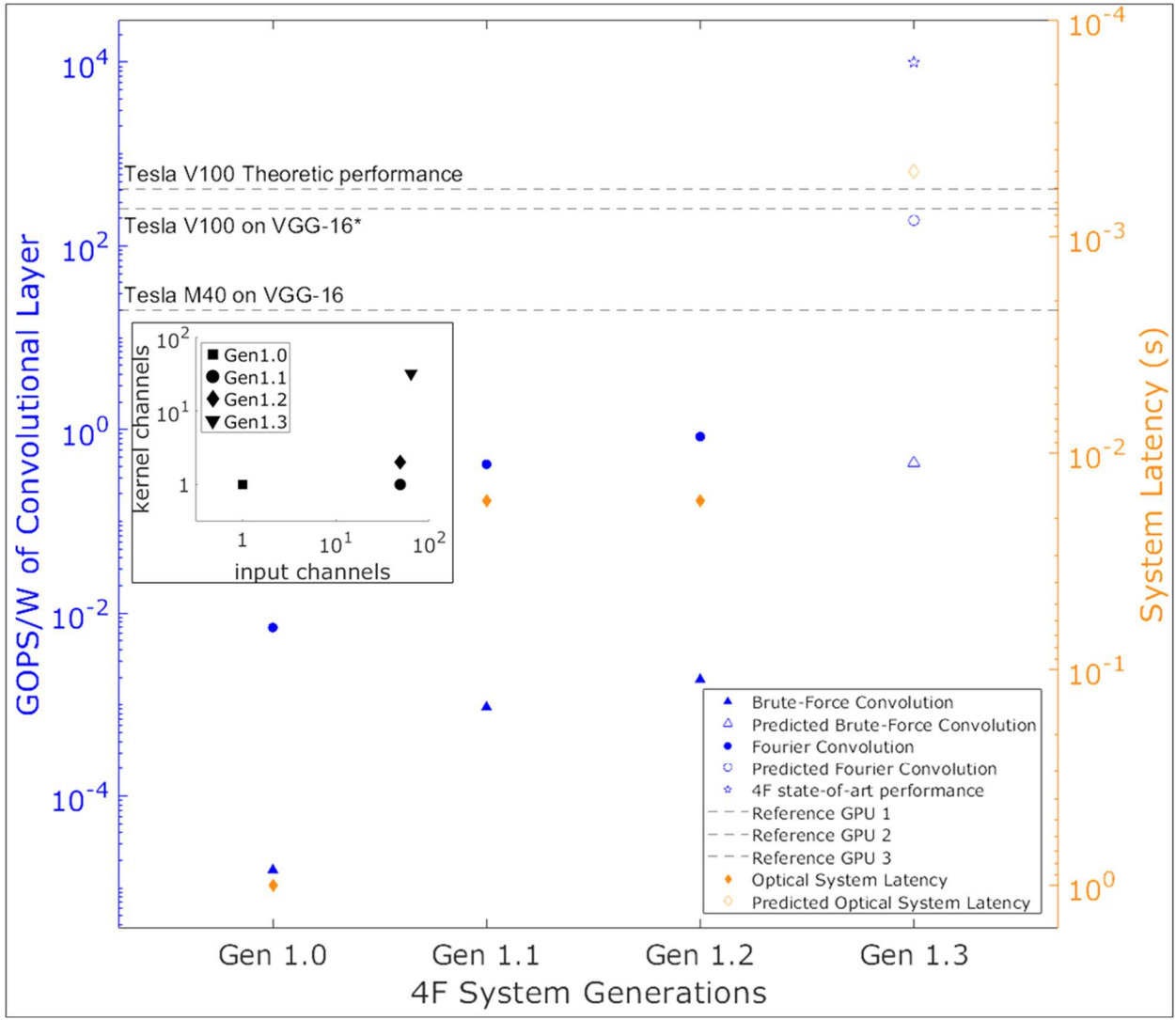

**Figure 4:** The equivalent computing power and the latency of the 4F system as compare to a GPU system. The generations evolve with their parallelism, 1×1, 49×1. 49×2, to 64×24, and the latency changes from 1Hz to 2kHz. For Gen.1.3 we plan to upgrade the focal length of the optical layer to 30mm to afford 32 kernel parallelization. We assumed practical alignment precision tolerance and optimal DMD resolution utilization. The GFLOPS/W is calculated with the digital equivalent convolution operations, while brute-force uses 3×3 kernel convolution operations and Fourier convolution results into two Fourier transform multiplication operations (Methods section). In this state-of-the-art system, we project 64×24 parallelism, full DMD resolution utilization which means we classify the 1920×1080 high resolution images instead of the 28×28 images from MNIST. **The Tesla V100 performance is extrapolated from the Tesla M40 (Supplementary S6.2) [37]. Tesla V100 uses tensor mixed-precision, M40 uses single-precision.

## 4. Conclusions

In this work we design and demonstrate an operation-parallelized (100× over state of art) high-throughput Fourier optic convolution accelerator. Our contributions include a multi-kernel convolution filtering paradigm enabled by harnessing higher order optical diffraction. This diffraction-enabled parallelism offers low-latency (10ms) and continuous operation and data streaming capability leading to high throughputs (~1 GOPS/W) which is capable of processing large-scale matrices (1000×1000). This approach offers systems capability exceeding 1 TOPS/W with 100's of microsecond inference latency. Additionally, Diffraction Convolutional Neural Network systems are capable of being part of an advanced all-optical neural network, which speeds-up to 10's of nanoseconds as input and output data being optical.

## 5. Methods

*5.1 Data preprocessing*

Unlike a digital FT, where the Fourier image is assumed to be the same size as the original image, the Fourier domain size in the optical Fourier system is constrained by the high, yet finite DMD resolution,

$$FT_{size} = \frac{\lambda f}{2\Delta} \quad (3)$$

where $\Delta$ is the pixel size of image, $\lambda$ is the carrier light wavelength, and $f$ is the focal distance of a lens. Therefore, we match the electronic (digital) to the optical FT for the training step (simulation) step (see Methods below for details). Our system's parameters are the focal length is 100mm, the wavelength is 450 nm, and the DMD pitch is 7.6 $\mu$m. Hence, in the experimental setup, the Fourier image size is 779 × 779 pixels. For consistency, in the simulation, we pad the original input image with zeros to the size of 779×779 and then use the digital FFT to simulate the optical FT.

However, the case above is still an idealized model of a physical D-CNN, therefore we upgrade the system to accomplish: 1) the optimization for coupling of modulated and carried signals in the Fourier domain; 2) the compensation for parasitic wavefront modulation due to the reflection at an angle at a DMD surface.

To address the first point, when building such an optical Fourier classifier one would observe blurriness in the back focal plane of the second lens. One trivial reason could be that the input image features are beyond the resolution capability of the setup. Interestingly, even if this criterion is met, in some cases the blur might be coming from a spatially large Fourier domain that exceeds the physical Fourier space (i.e. dimension) of the DMD#2. In this case, the Fourier domain in the back focal plane of L1 is too large to be effectively converted by L2 due to the finite physical size of a focal plane, see Figure. 1b, d. According to our empirical observations, the 3×3 superpixel applied to an input image results in the best performance of the classifier.

As for the second goal, the DMDs used for spatial amplitude modulation have a static $12^o$ tilt to each micromirror in the "off" state. These results in three effects: 1) distortion of the wavefront; 2) vertical compression of the image; 3) inability to position the entire image at a focal plane of L1 and L2. To resolve the first issue, we position the DMDs at 57º incidence angle, see Fig. 3, to account for this extra tilt. To account for the compression, we enlarge the input image horizontally by a factor of 1/cos 57º. Experimentally, the best performance has been achieved when enlarging by an integer factor of 2 due to both the DMD's binary modulation and the spatial quantization. There is no easy fix for the third issue. One way to mitigate it could be to consistently account for it in the simulation by point-by-point propagation modeling. In that case, the corresponding artifacts would have been accounted for on the level of the weights' generation for the optical convolutional layer. However, such an optical simulation poses substantial theoretical, numerical, and computational challenges. The results presented here do not account for this effect yet. However, the associated phase error is static and systematic, hence, potentially, possible to train for in an on-sight trainable D-CNN.

We analyze performance of this D-CNN classifier using CIFAR-10, Google quickdraw, and MNIST datasets. Custom data pre-processing procedures have been required to load the data into the classifier, including quantizing an 8-bit depth image into subsets of binary images, parallelizing the inputs, and adjusting the size of images. Importantly, due to using the DMDs for signal modulation, our setup can only operate on binary data, while ensuring a high system throughout. To convert the original grey level MNIST and Google quickdraw images into a 1-bit sequence we set a threshold at 80% for the maximum pixel value while the images maintain 90% structural similarity index. For CIFAR-10 8-bit RGB dataset, we quantize each image into 12 binary images by setting 4 evenly spaced thresholds over the full dynamic range. In this way, one ends up with 4 binary sequences per channel to load one CIFAR-10 image.

*5.2 Machine learning training methodology*

We train our model using the PyTorch platform and integrate the simulation model of the optical system into a custom Fourier convolution layer (Figure. 1c). All functions of the Fourier convolution layer are written using PyTorch's built-in functions so that PyTorch's autograd function can be directly applied for backpropagation during training. We train in the noise free mode and compensate for the noise during the second stage calibration process.

To detail the two-step training process; first, the fully electronic NN has been pre-trained to generate a set of Fourier kernels for a particular dataset. The Fourier domain kernel values, which are the weights of the convolutional layer, are initialized directly in the Fourier domain. The weights are trained with floating point precision, and binarization is applied in the forward pass of training to simulate DMD's binary modulation. Then, in the optical convolution layer, the system is fed with the same dataset of images loaded on DMD#1, and the pre-trained kernels loaded on DMD#2 in the Fourier domain, see Figure.1b, d. The output intensity distribution is registered with a camera to accomplish an optoelectronic domain conversion. Second stage is to train the weights in the fully connected layer with these camera outputs. Hence, the resulting network weights encompass a combination of Fourier kernel weights trained in step 1 and fully connected layer weights trained in step 2.

The Fourier convolution layer consists of 16 binary frequency-domain kernels, and the input size in case with, e.g., CIFAR10 is $32\times32$. The output size of the Fourier convolution layer after max-pooling is $16\times16\times16$. The output is fed into a fully connected layer with the size $4096\times256$, followed by the final output layer with the size $256\times10$ (or number of classes of the target dataset). During the training stage we binarize the kernel weights to be hardware compatible. The binarization scheme we used here is different from the common sign function where weight values are converted to -1 or +1. Since a DMD does not support negative values, we modified the sign function scheme; positive weights are replaced with 1 while negative weights are replaced with 0.

*5.3 Performance benchmarking*

To find the energy efficiency of our system we used the eq-s. (4-5) applied to the MNIST/Google Quickdraw dataset. To find input parallelism $i$ of the system we considered a 11×19 set of tiles to fill the DMD's space. This was determined by applying an expansion to 28 × 28 images, a 3 times expansion was used to account for superpixels on the DMD and 2 times was used for padding around each image. This brings the image size up to 168 × 168 DMD pixels where a set of 11 × 6 images fits on the 1920 × 1080 DMD. Note that the scaling of energy efficiency between this heterogeneous system and a GPU will change depending on the image size. Hence, it is best to consider image sizes which map directly to a CNN architecture's convolutional layer size to enable a direct comparison. Here VGG-16's 4th convolutional layer was used for comparisons(Supplementary S6.1) . The GOPS/W is the result of a Tesla M40 inferencing ImageNet using the VGG-16 architecture, a batch size of 128 was used to reach a utilization around 90% [37]. This figure was then used to extrapolate the GOPS/W for a Tesla V100 operating under the same conditions(Supplementary S6.2) .

First, the GOPS/W were calculated using the brute force convolution method with a kernel size of 3 x 3. This was done since VGG-16's 4[th] convolutional layer is calculated in this way rather than through the use of an FFT method (blue triangles, see Figure. 4). However, it is important to note that the FFT convolution method is closer to what the heterogeneous system is doing since the full convolution is done in the optical domain. Second, the energy efficiency of a Fourier convolution-based CNN was assessed (blue circles, see Figure. 4). One can easily see that the choice of a correct NN architecture can boost the optical accelerator performance by two orders of magnitude with minimal to no effort and/or design alteration. Hence, mimicking pre-existing electronic NN architectures is by far not an optimal approach in optical heterogeneous computing. With current NN architectures being optimized for electronic hardware, network architectural search, as applied to opto-electronic accelerators, is a field of study yet to flourish.

FFT:
$$\text{OPS/W} = \frac{[10 \cdot M \log((2i-1) \cdot M) \cdot N \log(N) + m \cdot n] \cdot i \cdot k \cdot f}{P} \quad (4)$$

Brute Force:
$$\text{OPS/W} = \frac{(M \cdot N \cdot m \cdot n) \cdot i \cdot f \cdot k}{P} \quad (5)$$

Where: $M \times N$ are the image size, $m \times n$ are the kernel size, $f$ is the system frequency, P is the power consumed by the system which is the summation of a camera(17W) and two DMDs(6.3W×2, see Supplementary S1), $i$ is the number of images tiled on the first DMD, $k$ is the kernel parallelization. The factor of 10 in eq. (4) comes from the FFT algorithm implementation complexity C×N×Log(N), where in this case C=5; and the additional factor of 2 comes from two FTs being performed in a single 4f-convolution. These formulas represent the energy efficiency of the heterogeneous system using the FFT and Brute force method for convolution. Since there is no direct analogue in an optical accelerator to the OP as defined in electronic, this formula considers the number of OPs that would be done by a GPU to perform a comparable computation [38]. In this way we are consider $i$ different convolutions where we have 2× operations for an FFT to represent the forward and inverse FT, and a dot product multiplication in the size of the kernel. These total operations are then multiplied by $k$ for parallelized kernels & multiplied by the frequency of the system resulting in the system's OPS. From there the energy efficiency is calculated by dividing the OPS by the Watts used by the optical components during computation. The I/O and memory energy are modeled using the access energy of DRAM and assuming that all the data is going directly to or coming from the DRAM. The resulting overall power consumption related to memory access is of the order of 100mJ, which is three orders of magnitude lower than the average power consumption of other devices, hence, it was neglected in calculations.

## Acknowledgements

This work was supported by the Office of Naval Research.